\documentclass[letterpaper, 10 pt, conference]{ieee/ieeeconf}  

\IEEEoverridecommandlockouts                              
\usepackage{graphics} 
\usepackage{epsfig} 
\usepackage{mathptmx} 
\usepackage{times} 
\usepackage{amsmath} 
\usepackage{amssymb}  
\usepackage[usenames, dvipsnames]{color}
\usepackage[linesnumbered,ruled,vlined]{algorithm2e}
\usepackage[caption=false,font=footnotesize]{subfig}
\usepackage[export]{adjustbox}
\usepackage{graphicx}
\usepackage{float}
\usepackage{multirow}
\usepackage{soul}

\definecolor{purple}{RGB}{149,33,246}

\newcommand{\YOON}[1] {
	\textcolor{red}{\bfseries{YOON: {#1}}}
}

\newcommand{\IK}[1] {
	\textcolor{blue}{\bfseries{IK: {#1}}}
}

\newcommand{\JW}[1] {
	\textcolor{purple}{\bfseries{JW: {#1}}}
}

\renewcommand{\baselinestretch}{1}

\title{\LARGE \bf
Diffraction-Aware Sound Localization for a Non-Line-of-Sight Source
}

\newcommand{\Skip}[1]{}
\renewcommand{\paragraph}[1]{{\bf {#1}}} 

\begin{document}

\author{Inkyu An$^{1}$, Doheon Lee$^{2}$, Jung-woo Choi$^{3}$, Dinesh Manocha$^{4}$, and Sung-eui
Yoon$^{5}$ \\
\thanks{$^{1}$I. An, $^{2}$D. Lee, and $^{5}$S. Yoon (Corresponding author) are
with the School of Computing, KAIST, Daejeon, South Korea;
	$^{3}$J. Choi is with the School of Electrical Engineering,
	KAIST;
	$^{4}$D. Manocha is with the Dept. of CS \& ECE, Univ. of Maryland at College Park, USA;
	{\tt\small \{inkyu.an, doheonlee, jwoo\}@kaist.ac.kr, dmanocha@gmail.com, sungeui@kaist.edu}}%
}

\maketitle
\thispagestyle{empty}
\pagestyle{empty}

\begin{abstract}
We present a novel sound  localization
algorithm for a non-line-of-sight (NLOS) sound source in indoor environments. Our approach exploits the diffraction properties of sound waves as they bend around a barrier or an obstacle in the scene.  
We combine a ray tracing based sound propagation algorithm with a Uniform Theory of Diffraction (UTD) model, which simulate bending effects by placing a virtual sound source on a wedge in the environment.
We precompute the wedges of a reconstructed mesh of an indoor scene and use them to generate diffraction acoustic rays to localize the 3D position of the source.
Our method identifies the convergence region of those generated acoustic
	rays as the estimated source position based on a particle filter.
We have evaluated our algorithm in multiple  scenarios consisting of  a static and dynamic NLOS sound source. In our tested cases, our approach can localize a source position with 
an average accuracy error, 0.7m,  measured by the L2 distance between estimated and actual source locations in a 7m$\times$7m$\times$3m  room. 
Furthermore, we observe 37\% to 130\% improvement in accuracy over a state-of-the-art  localization method that does not model diffraction effects, especially when a sound source is not visible to the robot.
%
\end{abstract}

\section{INTRODUCTION}
\label{sec:1}

\begin{figure}[t]
	\centering	
	\subfloat[A Non-Line-of-Sight (NLOS) moving source scene around an
	obstacle. Our method can localize its position using acoustic
	sensors and our diffraction-aware ray tracing.]
	{\includegraphics[width=0.8\columnwidth]{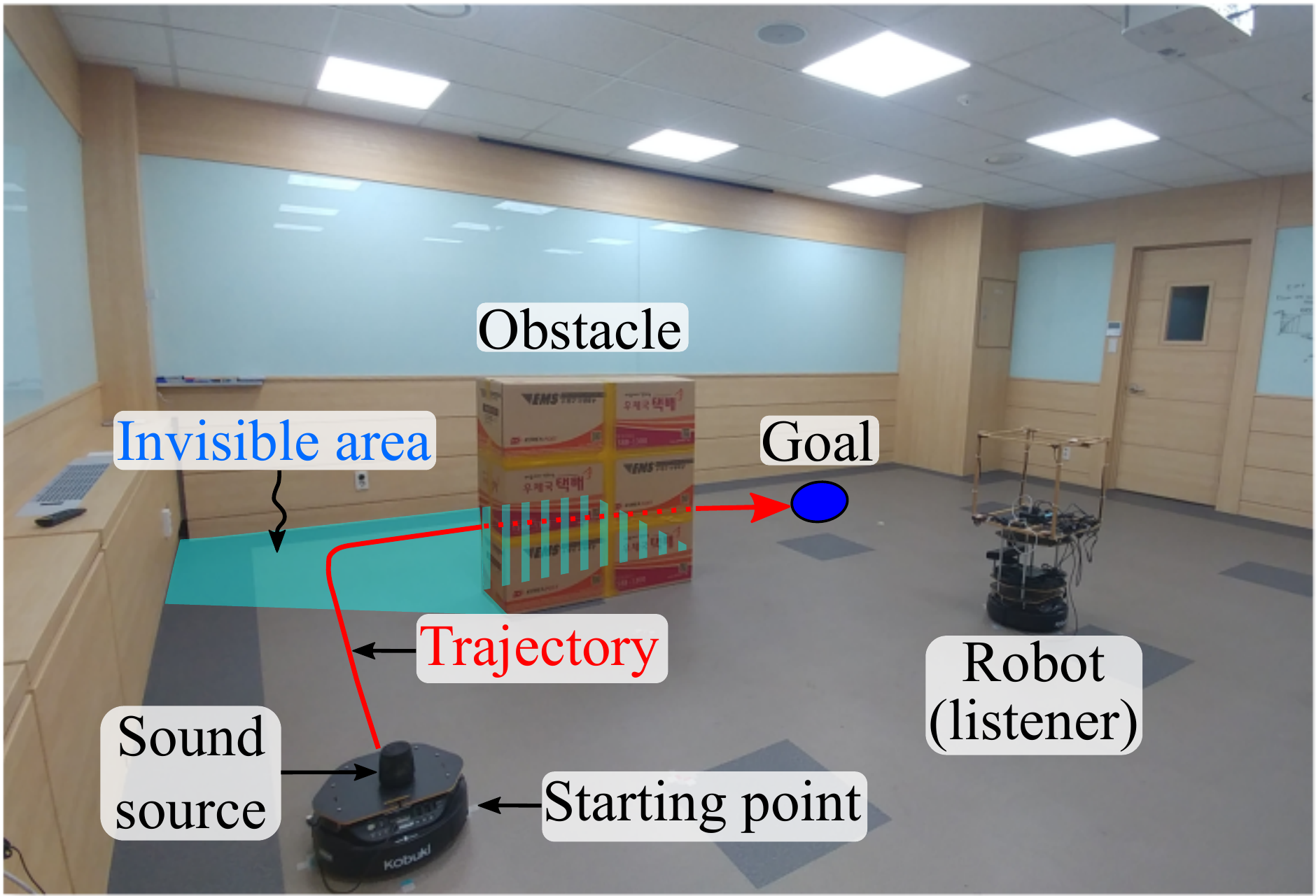}\label{fig:environment_moving_ss}}\\
	\subfloat[Accuracy errors, measured as the L2 distance between the estimated and actual 3D locations of a sound source, for the dynamic source. Our method models diffraction effects and improves the localization accuracy as compared to only modeling indirect reflections~\cite{an2018reflection}]
	{\includegraphics[width=0.8\columnwidth]{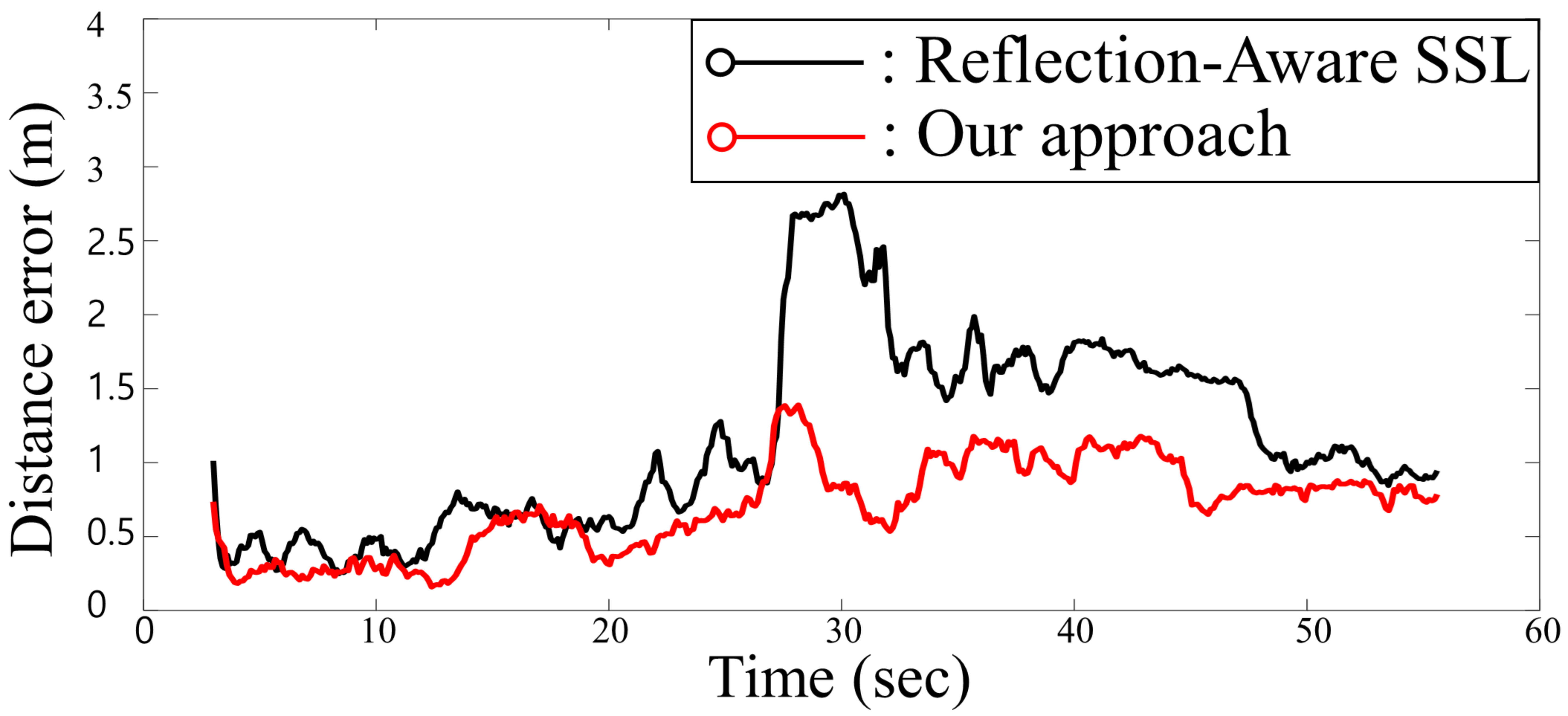}\label{fig:result_moving_ss}}
	\caption{ 
		These figures show the testing environment (7m by 7m with 3m height) (a) and the accuracy
		error of our method with the dynamically moving sound source
		(b).
		The source
		moves along the red trajectory, and the obstacle
		causes the invisible area for the dynamic source.
		Invisibility of the source occurs 
		from 27s to 48s, where our method maintains a high accuracy, while the prior method deteriorates due to the diffraction: 
		the average distance errors of our and the prior method are 0.95m and 1.83m.
	}
	\vspace{-1em}
	\label{fig:environment_resultGraph_moving_ss}
\end{figure}

As  mobile robots are increasingly used for different applications, there is considerable interest in developing new and improved methods for  localization. The main goal is to compute the current location of the robot with respect to its environment. Localization is a fundamental capability required by an autonomous robot, as the current location is used to guide the future movement or actions. We assume that a map of the environment is given 
and different sensors on the robot are used to estimate its position and orientation in the environment. 
Some of the commonly used sensors include GPS, CCD or depth cameras, acoustics, etc. In particular, there is considerable work on using acoustic sensors for localization, including sonar signal processing for underwater localization and microphone arrays for indoor and outdoor scenes. In particular, the recent use of smart microphones in commodity or IoT devices (e.g., Amazon Alexa) has triggered interest in better acoustic localization methods~\cite{douglas2017human,imran2016methodology},


The acoustic sensors use the properties of sound waves to compute the source location. As the sound waves are emitted from a source, they transmit through the media and either reach the listener or microphone locations as direct paths, or after undergoing different wave effects including reflections,  interference, diffraction,
scattering, etc. 
Some of the earliest work on sound source localization (SSL) makes use of the 
time difference of
arrival (TDOA) at the receiver~\cite{knapp1976generalized,he2018deep}. These methods only exploit the direct sound and its direction at the receiver, and do not take into account of reflections or other wave effects.  As a result, it does not provide sufficient accuracy for many applications. 
Other techniques have been proposed to localize the position under different constraints or sensors~\cite{sasaki2016probabilistic,su2017towards,misra2018droneears,an2018reflection}. This includes modeling of higher order specular reflections~\cite{an2018reflection} based on ray tracing and can model indirect sound effects. 

\begin{figure*}[t]
	\centering
	\includegraphics[width=2\columnwidth]{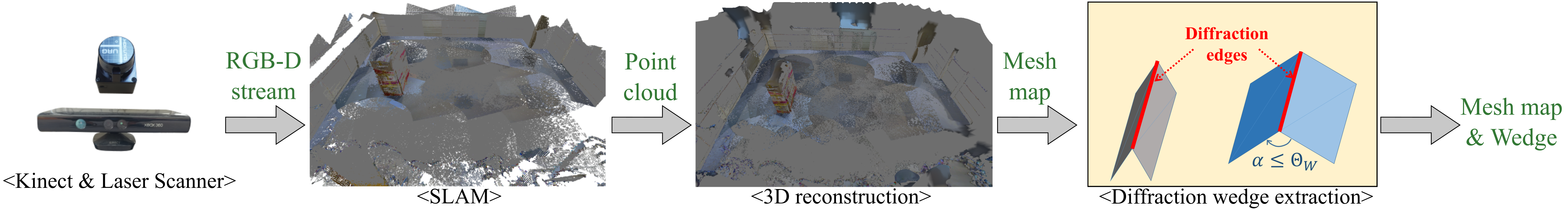}
	\vspace*{-0.3cm}		
	\caption{
		This figure shows our precomputation phase.
	We use SLAM to generate a
		point cloud of an indoor environment
		from the laser scanner and Kinect.  The point cloud is used to
		construct the mesh map via 3D reconstruction techniques.
		Wedges whose two neighboring triangles have angles larger 
		than $\theta_W$ and their edges are extracted from the mesh map to consider  
		diffraction effects at runtime for sound localization. 
	}
	\vspace{-1.0em}
	\label{fig:blockDiagram_precomputation}
\end{figure*}



In many scenarios, the sound source is not directly in line of sight of the listener (i.e. NLOS) and is occluded by obstacles. In such cases, there may not be much contribution in terms of direct sound, and simple methods based on TDOA may not work well. We need to model indirect sound effects and the most common methods are based on using ray-based geometric propagation paths. They assume the rectilinear propagation of sound waves and use ray tracing to compute higher order reflections. While they work well for high frequency sounds, but do not model many low-frequency phenomena like diffraction that is a type of scattering that occurs from obstacles whose size is of the same order of magnitude as the wavelength. In practice, diffraction is a fundamental mode of sound wave propagation and occurs frequently in building interior (e.g. source is behind an obstacle or hidden by walls).
These effects are more prominent for low-frequency sources, such as vowel sounds in human speech, 
industrial machinery, ventilation, air-conditioned units.

\paragraph{Main Results.} We present a novel sound localization algorithm that takes into diffraction effects, especially from non-light-of-sight or occluded sources. Our approach is built on a ray tracing framework and models diffraction using the Uniform Theory of Diffraction (UTD)~\cite{keller1962geometrical} along the wedges.
During the precomputation phase, we use SLAM and reconstruct a 3D triangular mesh for an 
indoor environment.
At runtime, we generate direct acoustic rays towards incoming sound directions as 
computed by TDOA. Once the acoustic ray hits the reconstructed mesh, we generate reflection
rays. Furthermore, when acoustic rays pass close enough to the edges of mesh wedges according to our diffraction-criterion, we also generate diffraction acoustic rays to model non-visible paths to include an incident sound direction that can be actually traveled
(Sec.~\ref{sec:4}). 
\Skip{
We consider not only the direct and reflection signals but only the diffraction
signals, and generate the primary, reflection, and diffraction acoustic rays in
the reconstructed environment (Sec.~\ref{sec:4}).
We suggest the notion of ~\textit{diffraction-condition} for determining the condition when diffraction rays are generated 
and explain how to create the diffraction ray on the wedges in Sec.~\ref{sec_diffraction}.
}
Finally, we estimate the source position by performing generated acoustic rays using ray convergence. 

We have evaluated our method in an indoor environment with three different scenarios, which include a stationary and a dynamically moving source along an obstacle that blocks the direct line-of-sight from the listener. 
In these cases, the diffracted acoustic waves are used to localize the position.
We combine our diffraction method with reflection-aware SSL algorithm~\cite{an2018reflection} and observe  
improvements from $1.22$m to $0.7$m on average and from $1.45$m to $0.79$m for the NLOS source.
Our algorithm can localize a source generating the clapping sound within $1.38$m as the worse error bound
in 
a room of dimension $7m \times 7m$ and $3$m height. 

\section{Related Work}
\label{sec:2}

In this section, we give a brief overview of prior work on sound source localization and sound propagation.

\paragraph{Sound source localization (SSL).}
Over the past two decades, many approaches 
have used time difference of arrival (TDOA) to localize sound sources.
Knapp 
\textit{et al.} presented a good estimation of the time difference using a generalized correlation between
a pair of microphone signals~\cite{knapp1976generalized}.
He \textit{et al.}~\cite{he2018deep} suggested a deep neural network-based 
source localization
algorithm in the azimuth direction for multiple sources.
This approach focused on estimating an incoming direction of a sound and 
did not localize the actual position of the source.

Recently, many techniques have been proposed to estimate the  location of a sound
source~\cite{sasaki2016probabilistic,su2017towards,misra2018droneears}.
Sasaki \textit{et al.}~\cite{sasaki2016probabilistic}
 and Su \textit{et al.}~\cite{su2017towards} presented 3D sound source localization
algorithms using a disk-shaped 
sound detector and a linear microphone array
such as Kinect and PS3 Eye.
Misra \textit{et al.}~\cite{misra2018droneears} suggested a robust localization method in
noisy environments
using a drone.
This approach requires the accumulation of steady acoustic signals at different positions, and thus cannot be applied to a transient sound event or to stationary sound detectors.

An \textit{et al.}~\cite{an2018reflection} presented a reflection-aware sound
source localization algorithm that used direct and reflected acoustic rays 
to estimate a 3D source position in indoor environments.
Our approach is based on this work and takes into account diffraction effects to considerably improve the accuracy.

\Skip{We found that its accuracy decreases in complex environments with 
many objects, mainly 
due to low-frequency effects, e.g. diffraction and diffuse. In this
work, our approach focuses on addressing this issue.
}


\paragraph{Interactive sound propagation.}
There is considerable work in acoustics and physically-based modeling to develop fast and accurate sound
simulators that can generate realistic sounds for computer-aided design and virtual environments.
Geometry acoustic (GA) techniques have been widely utilized to 
simulate sound propagations efficiently using  ray tracing techniques.  Because ray
tracing algorithms are based on the sound propagation model at high
frequencies, low-frequency wave effects like diffraction are 
modeled separately.

In addition, an estimation of the acoustic impulse response between the source and
the listener was performed using Monte Carlo path tracing~\cite{schissler2014high} or a hybrid combination of geometric and numeric methods techniques~\cite{yeh2013wave}.

\begin{figure*}[t]
	\centering
	\includegraphics[width=2\columnwidth]{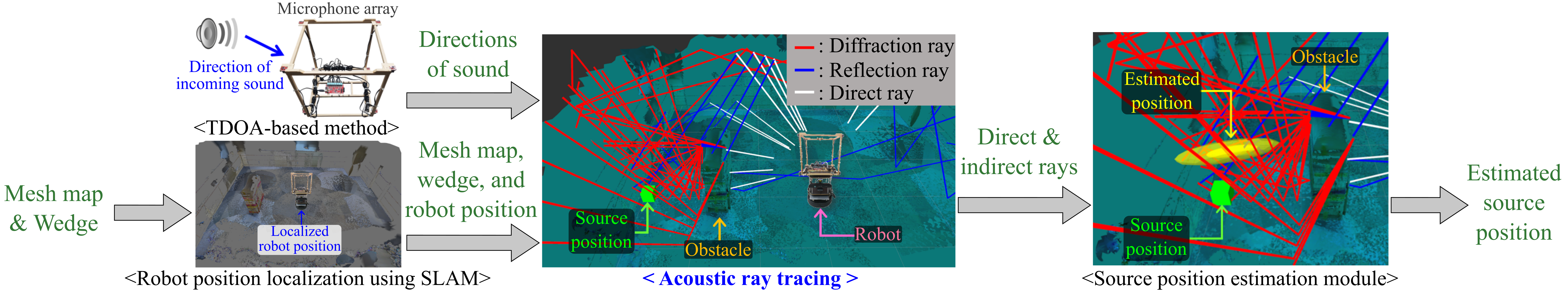}
	\vspace*{-0.3cm}		
	\caption{
		We show run-time computations using acoustic ray
		tracing with diffraction rays for sound source localization.
		The diffraction-aware acoustic ray tracing is highlighted in blue and our main contribution in this paper. 
        The source position estimation is performed by identifying ray convergence.
	}
	\vspace{-1em}
	\label{fig:blockDiagram_runtime}
\end{figure*}

Exact methods to model diffraction are based on directly solving the acoustic wave equation using numeric methods like boundary or finite element methods~\cite{teng1995new, martin2016hybrid} or the BTM
model~\cite{svensson1999analytic} and its extension to higher order diffraction
models~\cite{asheim2013integral}.
Commonly used techniques to model diffraction with geometric acoustic methods are based on two models:
the Uniform Theory of Diffraction (UTD)~\cite{kouyoumjian1974uniform} and the
Biot-Tolstoy-Medwin (BTM) model~\cite{svensson1999analytic}.  The BTM model is
an accurate diffraction formulation that computes an integral of the
diffracted sound along the finite edges in the time
domain~\cite{asheim2013integral,martin2016hybrid,antani2012efficient}. 
In practice, the BTM model is more accurate, but is limited to non-interactive applications.
The UTD model approximates an infinite wedge as a secondary source of
diffracted sounds, which can be reflected and diffracted again before reaching
the listener.  UTD based approaches have been effective for
many real-time sound generation applications, especially in complex
environments with occluding
objects~\cite{tsingos2001modeling,schissler2014high}. Our approach is motivated by these real-time simulation and proposes a real-time source localization algorithm using UTD.


~

\section{Diffraction-Aware SSL}
\label{sec:4}

We present our diffraction-aware SSL based on acoustic ray tracing, starting
with giving its overview.

\subsection{Overview}

\paragraph{Precomputation.} Given an indoor scene, we reconstruct a 3D model as part of the precomputation.
We use a Kinect and a laser scanner to capture a 3D point cloud representation of the indoor scene. As shown in
Fig.~\ref{fig:blockDiagram_precomputation}, the point cloud capturing the
indoor geometry information is generated by the SLAM module from raw depth
data and an RGB-D stream collected by the laser scanner and Kinect. Next, we
reconstruct a 3D mesh map via the generated point cloud. We also extract wedges
from the mesh that have angle, between two neighboring triangles, smaller than the threshold, $\Theta_W$.
The reconstructed 3D mesh map and the wedges on it are used for our diffraction
method at runtime.

\paragraph{Runtime Algorithm.} We provide an overview of our runtime algorithm as it performs acoustic ray tracing and sound
source localization in Fig.~\ref{fig:blockDiagram_runtime}.
Inputs to our runtime algorithm are the audio stream collected by
the microphone array, the mesh map reconstructed in the precomputation, and the
robot position
localized by the SLAM algorithm. Our goal is find the 3D position of the sound source in the environment.
Based on those inputs, we perform acoustic ray tracing supporting direct,
reflection, and diffraction effects by generating various acoustic rays
(\ref{sec:4_1}).  The source position is computed by estimating the convergence region of the acoustic rays 
(\ref{sec:Estimating_source}).  Our novel component, acoustic ray
tracing with diffraction rays,
is highlighted by the blue font in Fig.~\ref{fig:blockDiagram_runtime}.

\subsection{Acoustic Ray Tracing}
\label{sec:4_1}

In this section, we explain how our acoustic ray tracing technique generates
direct, reflection, and diffraction rays.

At runtime, we first collect the directions of the incoming sound signals from the
TDOA algorithm~\cite{valin2007robust}. 
For each incoming direction, we generate a primary acoustic ray in the backward 
direction; as a result, we perform acoustic ray tracing
in a backward manner. 
At this stage, we cannot determine whether the incoming signal is generated by one of the states: direct propagation, reflection, or diffraction.
We can determine the actual states of these primary acoustic rays 
  while performing 
backward acoustic ray tracing.  Nonetheless, we denote this primary ray as the
direct acoustic ray since the primary ray is a direct ray from 
the listener's perspective.


\Skip{
First, we generate the direct acoustic ray $r_n(l)$ using the collected
direction $\hat{v}_n$ of the $n$-th incoming sound signal:
to the inverse direction $\hat{d}_n^0$ of the direction $\hat{v}_n$ of the
$n$-th incoming sound signal, i.e., $\hat{d}_n^0=-\hat{v}_n$:

\begin{equation}
\begin{aligned}
r_n^0(l) = \hat{d}_n^0 \cdot l + \dot{o},
\end{aligned}
\end{equation}
}

We represent a primary acoustic ray as $r_n^0$ 
for the $n$-th incoming sound direction. 
Its superscript denotes the order of the acoustic path,
where the 0-th order denotes the direct path from the listener. 
We also generate a (backward) reflection ray once an acoustic ray intersects with the
scene information under the assumption that the intersected material mainly consists of specular materials~\cite{an2018reflection}.  The main difference from the prior
method~\cite{an2018reflection} is that we use a mesh-based representation, while the prior method used a voxel-based
octree representation for intersection tests.
This mesh is computed during precomputation and we use the triangle normals to perform the reflections. As a
result, for the $n$-th incoming sound direction, we recursively generate reflection rays with increasing orders, encoded by a ray path that is  defined  by
$R_n = [r_n^0,r_n^1,...]$.
The order of rays increases as we perform more reflection and diffraction.

\Skip{
	The whole process of
ray-tracing for reflected rays are based on the reflection-aware SSL
method.  As shown in
Fig.~\ref{fig:generating_reflected_ray}, the initial acoustic ray $r_n^0$ of
the $n$-th direction $\hat{d}_n^0$ is propagated to the free space, and the
specular reflection occurs when the initial acoustic ray hits the mesh 1; all
meshs are generated at the precomputation.  It is quite easy to compute the
outgoing reflected ray $r_n^1$ because we already know normal vectors of all
meshs.}

\subsection{Handling Diffraction with Ray Tracing}
\label{sec_diffraction}

We now explain our algorithm to model the diffraction effects efficiently within acoustic ray
tracing to localize the sound source.   
Since our goal is to achieve fast performance in localizing the sound source, we use the formulation based on Uniform Theory of Diffraction
(UTD)~\cite{kouyoumjian1974uniform}.
The incoming sounds collected by the microphone array consist of contributions from different effects in the environment, including reflections and diffractions.

Edge diffraction occurs when an acoustic wave hits the edge of a 
wedge. In the context of acoustic ray tracing, when an acoustic ray hits an edge of a wedge between two neighboring triangles, the diffracted signal
propagates into all possible directions from that edge.  The
UTD model assumes that the point on the edge causing the 
diffraction effect is an imaginary source generating the
spherical wave~\cite{kouyoumjian1974uniform}.
%

\begin{figure}[tp]
	\centering
	\subfloat[Generating diffraction rays]{\includegraphics[width=0.75\columnwidth]{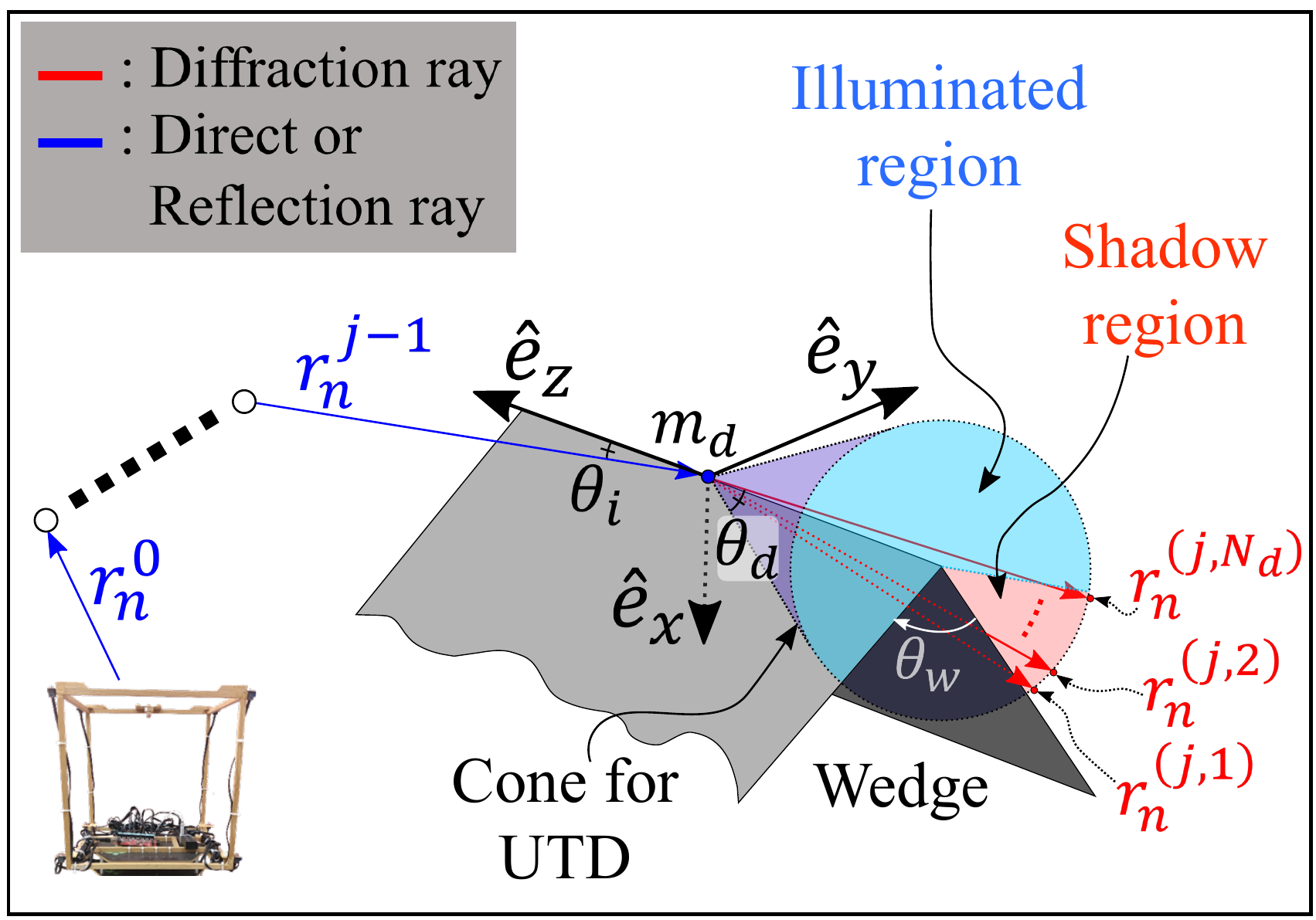}\label{fig:generating_diffracted_ray}} \
	\subfloat[Computing outgoing directions of diffraction rays.]{\includegraphics[width=0.7\columnwidth]{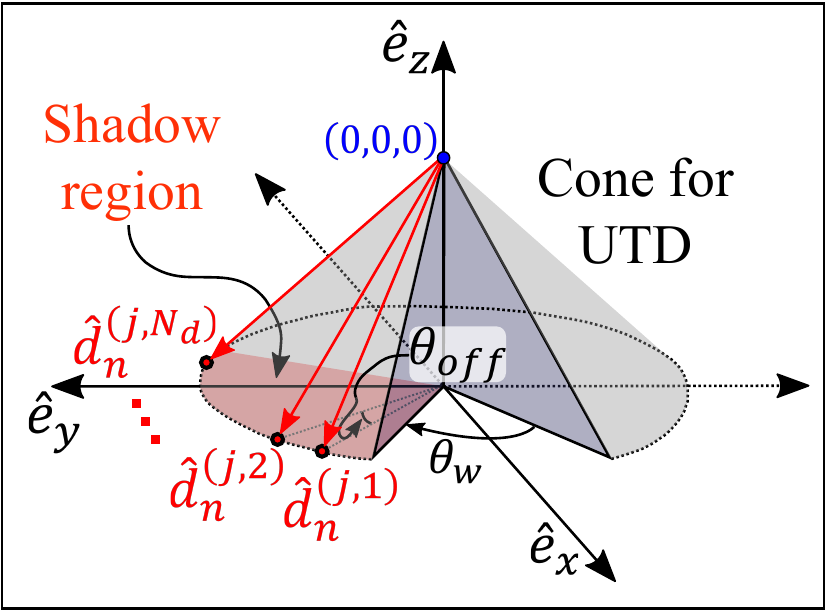}\label{fig:generating_diffracted_ray_local}}\qquad
	\caption{
		This figure illustrates our acoustic ray tracing method for handling
		the diffraction effect. 
(a) Suppose that we have an acoustic ray $r_n^{j-1}$ 
	satisfying
the diffraction condition, hitting or passing near the edge of a
	wedge.  We then generate $N_d$ diffraction rays covering the 
	possible incoming directions (especially, in the shadow region) of rays
	that cause the diffraction. 
	\Skip{
	After the outgoing
		unit vectors are converged to the global coordinate that is
		real environment, the $N_d$ number of diffraction rays are
		generated to a shadow region along a surface of a cone that has
		a same angle with an incident angle $\theta_i$.
	}
	(b)
	An outgoing unit vector, $\hat{d}_n^{(j,p)}$, of a $p$-th diffraction
ray is computed on local coordinates ($\hat{e}_x,\hat{e}_y,\hat{e}_z$), and
used after the transformation to
	the environment in runtime, where $\hat{e}_z$ fits on the edge of the
wedge and $\hat{e}_x$ is set half-way between two triangles of the wedge.
	}
	\vspace{-1em}
	\label{fig:acousticRayTracing}
\end{figure}

In order to solve the problem of localizing the sound source, we simulate the
process of backward ray tracing.  Suppose that an $n$-th incoming sound direction 
denoted by the ray $r_n^{j-1}$ is generated
by the diffraction effect at an edge. In an ideal case, 
the  incoming ray will hit the edge of a wedge and generate 
the diffraction acoustic ray $r_n^{j}$,
 as shown in Fig.~\ref{fig:generating_diffracted_ray}; in the figure of (a), $r_n^{(j,\cdot)}$ is shown.
\Skip{
one of
diffracted signals from the edge propagates to the inverse direction, $-XX$, of
the $n$-th incoming sound signal.
}
It is important to note that there can be an infinite number of incident rays
generating diffractions at the edge.
Unfortunately, it is not easy to link the incident direction exactly to the edge generating the diffraction.
Therefore, we generate a set of $N_d$ different diffraction rays in a backward manner that covers
the possible incident directions to the edge based on the UTD model.
This set is generated based on an assumption that  one of those generated rays might
have the actual incident direction causing the diffraction.  When there are sufficient acoustic rays, including the primary, reflection, and
diffraction rays, it is highly likely that those rays will pass through or near
to the sound source location; 
we choose a proper value of $N_d$ by analyzing diffraction rays (Sec.~\ref{sec:result_discussion}).

\Skip{
Since the diffraction effect mainly occurs at the wedges of the mesh, we
generate secondary, diffraction rays on those wedges of the mesh, once the
primary or reflection acoustic rays intersect or pass by those wedges.
\YOON{this sen. is unclear..}\IK{Addressed. modified this sentence.}
}

This explanation begins with the ideal case, where the 
acoustic ray $r_n^{j-1}$ hits the edge of a wedge.
Because our algorithm works on the real environment containing various types of
errors from sensor noises and resolution errors from the TDOA method, it is rare 
that an acoustic ray intersects an edge exactly.

\begin{figure}[t]
	\centering
	\includegraphics[width=0.8\columnwidth]{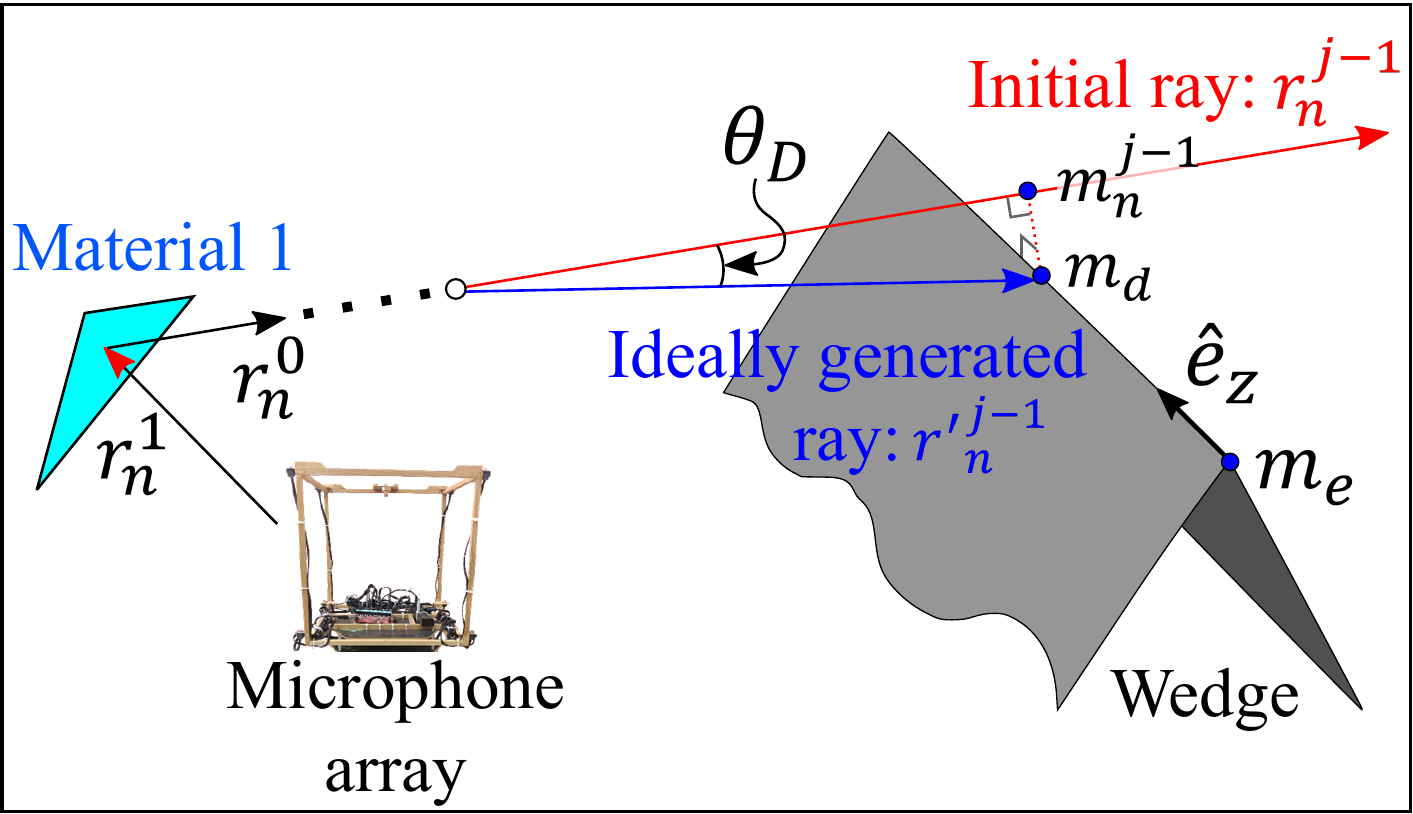}
	\caption{
	This figure shows the diffraction condition.  When a ray
	$r_n^{j-1}$
passes
close to an edge of a wedge, we consider the ray to be generated by the
	edge
diffraction.
We measure the angle $\theta_D$ between the ray and its ideal
generated ray that hits the edge exactly, for checking our diffraction condition.
	}
	\vspace{-1em}
	\label{fig:computing_diffractability}
\end{figure}

In order to support various cases that arise in real environments, we propose using the 
notion of \textit{diffraction-condition} between a ray and a wedge.
The diffraction-condition simply measures how close the ray
$r_n^{j-1}$
passes to an edge of the wedge.
Specifically, 
we define the \textit{diffractability} $v_d$ according to the angle $\theta_D$
between the  acoustic ray and its ideally generated ray for the diffraction with
the  wedge: i.e. $v_d = \cos(\theta_D)$, where the $\cos$ function is used to normalize the angle $\theta_D$ (Fig.~\ref{fig:computing_diffractability}).

Given an acoustic ray $r_n^{j-1}$, we define its ideally generated ray
${r'}_n^{j-1}$ as the projected ray of $r_n^{j-1}$ on the edge of the wedge
where the end point $m_d$ of ${r'}_n^{j-1}$ is on that edge (refer to the 
geometric illustration on Fig.~\ref{fig:computing_diffractability}).  The point
$m_d$ is located at the position closest to the point
$m_n^{j-1}$ of the input ray $r_n^{j-1}$; due to the page limit, we do not show its detailed derivation, but it can be defined based on our high-level description.

\Skip{ 
As shown in Fig.~\ref{fig:computing_diffractability}, given the unit vector
$\hat{d}_n^{j-1}$ of the propagation direction of $r_n^{j-1}$, the point
$m_d^{j-1}$ which is the starting point of $r_n^{j-1}$, 
and the point $m_d$ which is the end of that edge, 
we can compute the unit vector $\hat{d'}_n^{j-1}$ of the direction of $\hat{r'}_n^{j-1}$ from the two points, $m_d$ and $m_n^{j-1}$, which are the starting and end points of that ray, if we know the point $m_d$:
We construct the equations for two points, $m_d=m_e+\alpha \cdot \hat{e}_z$ and $m_n^{j-1}=m_d^{j-1}+ \beta \cdot \hat{d}_n^{j-1}$, where $\alpha$ and $\beta$ are the variables.
Then, we can obtain the answer for the variables, $\alpha$ and $\beta$, by solving a below equation which is 
suggested from the definition of the ideally generated ray, the vector $\hat{\gamma}$ of the shortest distance should be orthogonal 
Where $\hat{\gamma}$ denotes the vector from $m_d$ to $m_n^{j-1}$.

\begin{equation}
\begin{aligned}
\begin{split}
\hat{\gamma} =& [m_n^{j-1}-m_d]^T, \\
\hat{\gamma}^T \cdot \hat{d}_n^{j-1}& = 0, \
\hat{\gamma}^T \cdot \hat{e}_z = 0.
\end{split}
\end{aligned}
\end{equation}

}

If the diffractabilty $v_d$ is larger than a threshold value, e.g., $0.95$ in our
tests, our algorithm determines that the acoustic ray is generated from 
the diffraction at the wedge, and we thus generate the secondary,
diffraction ray at the wedge in the backward manner.

\Skip{
when an acoustic ray is traced, we compute the
\textit{diffractability} corresponding to the relation between the observation
ray and the ground truth ray inynd
generate the diffraction rays where that \textit{diffractability} satisfies the
diffraction condition.  
}

We now present how to generate the diffraction rays when the acoustic ray
satisfies the diffraction-condition.  The diffraction rays are generated
along the surface of the cone (Fig.~\ref{fig:generating_diffracted_ray})
because the UTD model is based on the principle of Fermat~\cite{keller1962geometrical}: the ray follows the
shortest path from the source to the listener.  
The surface of the cone for the UTD model contains every set of the shortest paths.
When the acoustic ray $r_n^{j-1}$ satisfies the diffraction-condition, we compute outgoing directions for those diffraction rays.
Those directions are the unit vectors generated on that cone and can be
computed on a local domain as shown in Fig.~\ref{fig:generating_diffracted_ray_local}:

\begin{equation}
\begin{split}
\hat{d}_n^{(j,p)} = \begin{bmatrix}
\cos{({\theta_w}/2 + p \cdot \theta_{off})} \sin{\theta_d} \\
\sin({\theta_w}/2 + p \cdot \theta_{off})\sin{\theta_d} \\
-\cos{\theta_d}
\end{bmatrix},
\end{split}
\label{eq:computing_unit_vector}
\end{equation}
where $\hat{d}_n^{(j,p)}$ denotes the outgoing unit vector of a $p$-th
diffraction ray among $N_d$ different diffraction rays, $\theta_w$ is the angle
between two triangles of the wedge, $\theta_d$ is the angle of the cone that is
same as the angle between the outgoing diffraction rays and the edge on the
wedge, 
and $\theta_{off}$ is the offset angle between two sequential diffraction rays,
i.e. $\hat{d}_n^{(j,p)}$and $\hat{d}_n^{(j,p+1)}$, on the bottom circle of the
cone.

Given a hit point $m_d$ by an acoustic ray
$r_n^{j-1}$ on the wedge, we transform the outgoing directions 
in the
local space to the world space by aligning their coordinates 
($\hat{e}_x,\hat{e}_y,\hat{e}_z$).
Based on those transformed outgoing directions, we then compute the outgoing
diffraction rays, $\bar{r}_n^{(j)}=\{r_n^{(j,1)},...,r_n^{(j,N_d)}\}$, starting
from the hit point $m_d$.

~\Skip{
Fig.~\ref{fig:generating_diffracted_ray}, 
Given a hit point $m_d$ by the incident ray $r_n^{j-1}$ on the
wedge,
diffraction rays
$\bar{r}_n^{(j)}=\{r_n^{(j,1)},...,r_n^{(j,N_d)}\}$ are propagated along the
surface of the cone \YOON{define it}, which has the same angle with the incident angle
$\theta_d$,  where the total number of diffracted rays is controled by the parameter
$N_d$.
}

In order to accelerate the process, we only generate the diffraction rays in
the shadow region, which is defined by the wedge; 
the rest of the shadow region is called the illuminated region. We use this process because
covering only the shadow region but not the illuminated region generates minor errors 
for a simulation of the sound propagation~\cite{tsingos2001modeling}. 

Given the new diffraction rays, we apply our algorithm recursively and generate another
order of reflection and diffraction rays. Given the $n$-th incoming direction signal, we generate acoustic rays, including direct, reflection, and diffraction rays
and maintain the ray paths $R_n$ in a tree data structure. The root of this tree 
represents the direct acoustic ray, starting from the microphones. 
The depth of the tree denotes the order of its associated ray. Note that we generate one child and $N_d$ children for
handling reflection and diffraction effects, respectively.

~

\subsection{Estimating the Source Position}
\label{sec:Estimating_source}

We explain our method used to localize the sound source position using
acoustic rays.  
Our estimation is based on Monte-Carlo localization (MCL), also known as the
particle filter~\cite{an2018reflection}. 
Our estimation process assumes that
there is a single sound source in the environment, which causes a high probability that
all those acoustic ray paths
pass  near that source;
Handling multiple targets using a particle filter has been also 
studied~\cite{okuma2004boosted}.
In other words, the acoustic rays converge in a region located close to the source, 
and our estimation aims to identify such a convergence region out of all the generated rays. 

The MCL approach generates initial particles in the space as an approximation to the
source locations. It allocates higher weights to particles that are closer to
acoustic rays and re-samples the particles to get more particles in
regions with higher weights~\cite{an2018reflection}. 
Specifically, we adopt the generalized variance, which is a one-dimensional measure for multi-dimensional scatter data, 
to see whether particles have converged.
When the generalized variance is less than a threshold (e.g.,
$\sigma_c=0.5$),
we treat that a sound occurs and the mean position of those particles 
as the estimated
sound source position.


\Skip{
The MCL approach consists of three parts: sampling, likelihood computation, and
resampling.  where the particle locations mean the hypothetical location of the
source, the concept is that particles are converged in the high likelihood
region by using the resampling process after the particles are sampled based on
the random distribution from the previous particles.
Since we design that the particles have the higher likelihood as close as the acoustic ray using the distance weight, 
the particles should have the highest weight in the convergence region of acoustic rays.
We determine the middle point of the convergence region as the estimated
position when the generalized variance is less than the threshold value (0.5).
}

~

\section{RESULTS and DISCUSSION}
\label{sec:result_discussion}

In this section, we describe our setup consisting of a robot with microphones and testing
environments, and highlight the performance 
of our approach.
The hardware platform is based on Turtlebot2 with a 2D laser scanner, Kinect, a
computer with an Intel i7 process, and a microphone array, which is an embedded
system for streaming multi-channel audios~\cite{briere2008embedded}, consisting
of eight microphones.  For all the computations, we use a single core, and
perform our estimation every 200ms, supporting five different estimations in
one second.

\Skip{
 that  does not have enough quality to be
immediately utilized for our acoustic ray-tracing algorithm,
we manually compensate meshes using a CAD software tool for the obstacles and filling the holes on the walls, bottom, and ceiling.
\JW{ Can't understand the meaning of the sentence. Do you need to compensate the meshes, and will it be explained later in this section? or the compensation is done but the way to do that is omitted because it's too simple? Also, there are too many typos in the manuscript.}~\IK{Addressed; we used a CAD tool for compensating the meshes manually..}
}

\begin{figure}[b]
	\centering
	\includegraphics[width=0.7\columnwidth]{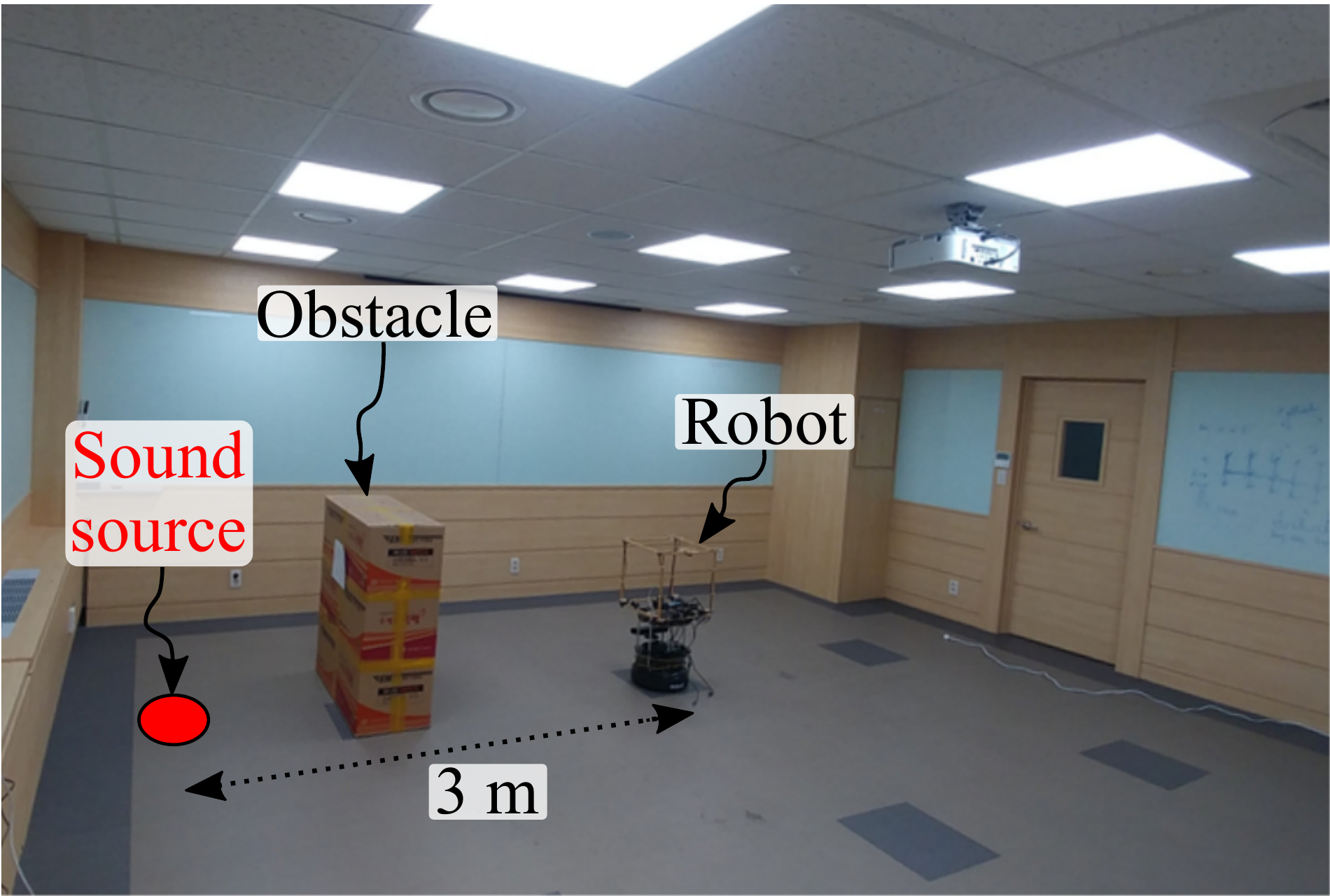}
	\vspace{-0.5em}
	\caption{
		The evaluation environment for the static sound source.
		Direct paths from the sound source to the listener are  blocked
by the obstacle. We use our diffraction-based algorithm for localization.
	}
	\label{fig:environment_stationary_ss}
\end{figure}

~

\paragraph{Benchmarks.}
We have evaluated our method in indoor environments containing a box-shaped object that
blocks direct paths from the sound to the listener. We use two scenarios: 
a stationary sound source and a moving  source.  As shown in
Fig.~\ref{fig:environment_stationary_ss}, we place an obstacle between the
robot and the stationary sound source, such that the source is not in the
direct line-of-sight  of the robot (i.e. NLOS source).
We use another testing environment with a source moving along the red
trajectory, as shown in Fig.~\ref{fig:environment_moving_ss}.
These two scenarios are tested on the same room that size is $7m \times 7m$ and $3$m height.

During the precomputation phase, 
we perform SLAM and reconstruct a mesh of the testing environment. We ensure that the resulting mesh has no holes using the MeshLab package.

\begin{figure}[t]
	\centering
		\subfloat[Stationary source (clapping).
		]{\includegraphics[width=0.9\columnwidth]{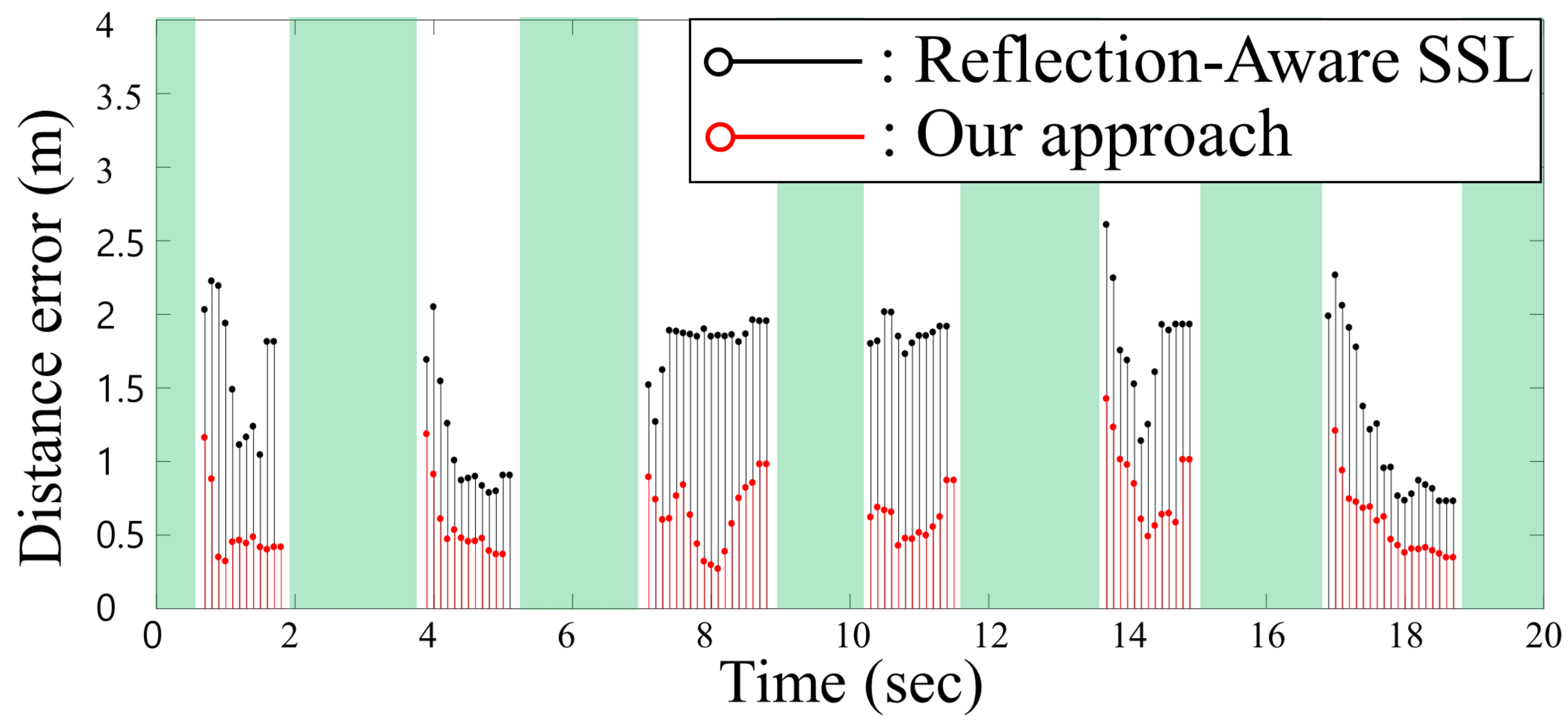}\label{fig:result_stationary_clapping_w_obs}}\\
		\vspace{-0.7em}
	\subfloat[Stationary source (male speech).]{\includegraphics[width=0.9\columnwidth]{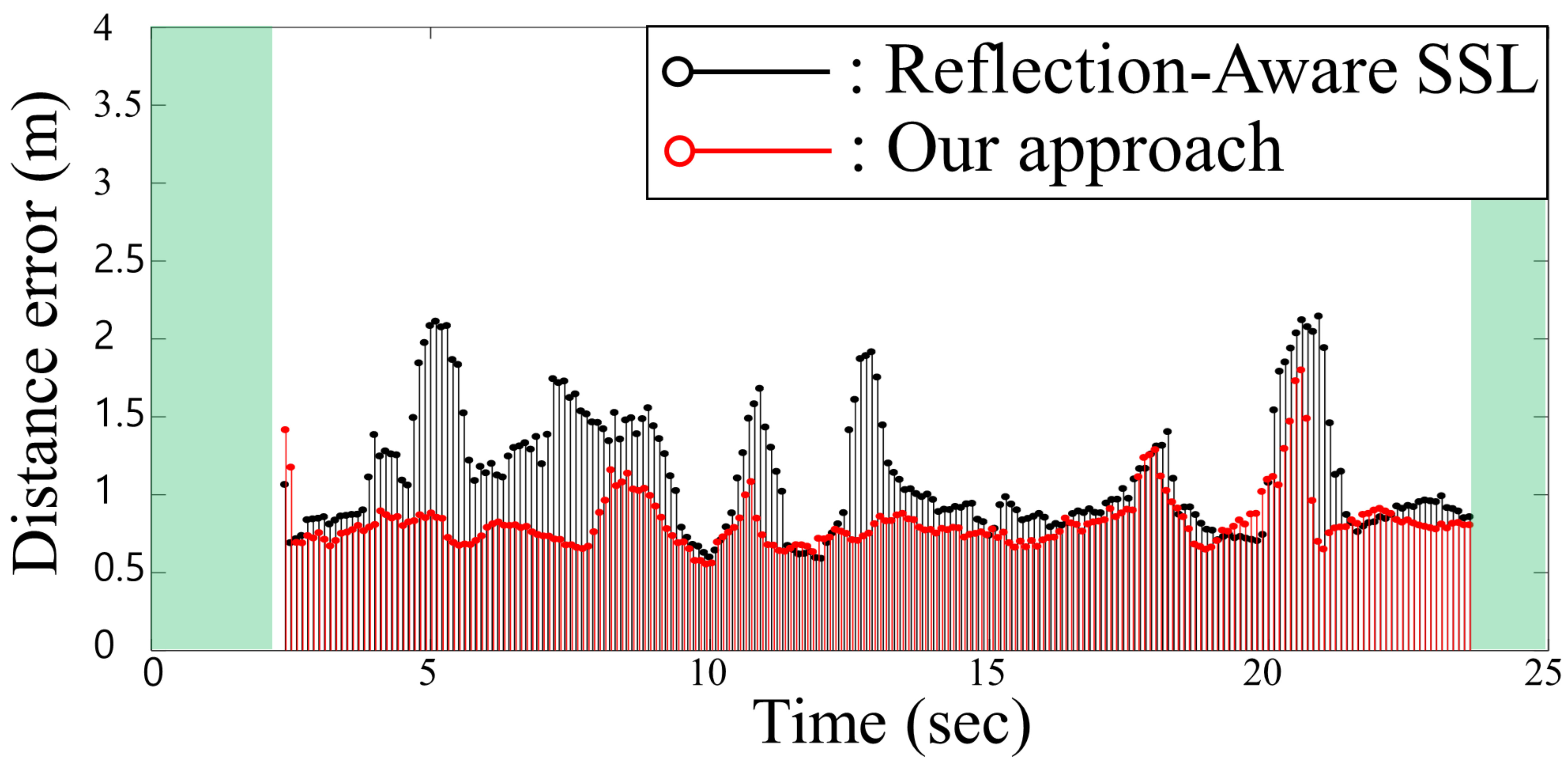}\label{fig:result_stationary_voice_w_obs}}
	\caption{
		These graphs compare the localization distance errors of our method
		with the prior, reflection-aware SSL method~\cite{an2018reflection} with the clapping
		sound source (a) and male speech signal source (b); green regions indicate no sound in that period.
		The average distance errors of RA-SSL and our method are 1.4m and 0.6m in (a), 
		and 1.12m and 0.82m in (b), respectively. The use of diffraction considerably reduces the localizatoin errors. 
		\Skip{
		the results, which are the distance errors between estimated
		source position and the actual source position, \JW{(what
		results? be specific. )}\IK{Addressed} of the stationary
		sources.  Black plots:  prior work (Reflection-aware SSL), red
		plots: our approach.
		(a) shows the distance errors of the clapping sound source, where the green background 
		indicates the intervals of no sound event.
		(b) shows the distance errors 
		in case of the continuous male's speech signal.		
	}
	}
	\vspace{-1em}
	\label{fig:resultGraph_stationary_ss}
\end{figure}

\paragraph{Stationary sound source with an obstacle.}
We evaluate the accuracy by computing the L2 distance errors between the
positions estimated by our method and the ground-truth positions.
We use two types of sound signals: the clapping sound and male speech, where 
male speech has more low-frequency components than the clapping sound (dominant frequency range of the clapping sound: 2k$\sim$2.5kHz, and of male speech: 0.1k$\sim$0.7kHz). 

We compare the accuracy of our approach with that of 
Reflection-Aware SSL (RA-SSL)~\cite{an2018reflection}, which models direct sound and indirect reflections, but no diffraction.
For the stationary source producing clapping sound
(Fig.~\ref{fig:result_stationary_clapping_w_obs}), the average distance errors
of the RA-SSL and our method are $1.4$m and $0.6$m, respectively.
There are configurations of the sound source that are not visible to the microphone (NLOS). In this case, we observe $130\%$ accuracy by modeling these diffraction rays. 

Fig.~\ref{fig:result_stationary_voice_w_obs} shows the localization accuracy
for the male speech signal, which has more low-frequency components.  The measured
distance errors are, on average, $1.12$m for RA-SSL and $0.82$m for our approach. 
While we also observe meaningful improvement, it is less than we see with the clapping sound.
Our method supports diffraction, but diffuse reflection is not yet supported.
Given the many low-frequency components of male speech, we observe that it is important to 
support diffuse reflection in addition to diffraction.
Nonetheless, by modeling diffraction for the male speech, we observe meaningful improvement (37\%
on average) in localization accuracy.
\Skip{
The reason why the accuracy decreases as compared with the clapping sound case
is that the number of specular reflection rays is decreased; low-frequency
sounds tend to produce more diffuse reflections, which are not considered in
this paper, rather than specular reflections.  
}
\Skip{\JW{(Please check your expressions;
diffuse reflections are different from diffraction. Which one is
correct?)}\IK{Addressed; I wanted to explain that the reason why the accuracy
decreases is that we didn't consider the diffuse reflections.}
}


\paragraph{Moving sound source around an obstacle.}
We also evaluated our algorithm on a  more challenging environment that contains a sound source (clapping 
sound) moving
along the red trajectory shown in Fig~\ref{fig:environment_moving_ss}.  Its accuracy graphs
are presented in Fig.~\ref{fig:result_moving_ss}; the
average distance errors of the RA-SSL and ours are $1.15$m and $0.7$m,
respectively, indicating a 64\% improvement in accuracy using our localization algorithm.
It is interesting that, 
when the dynamic source
is in the area corresponding to these time values (27s $\sim$ 48s), which are NLOS with respect to the robot,
the average distance errors of the
RA-SSL and our method 
are $1.83$m and $0.95$m, respectively, indicating a 92\% improvement.
This clearly demonstrates the benefits of our method in terms of localization.

Overall, we achieved 130\%, 37\%, and 64\% improvement, resulting in 77\% average improvement,
on the stationary source with a clapping sound, the stationary source with male
speech, and the dynamic source, respectively, compared with the prior method RA-SSL~\cite{an2018reflection}.
The summary of the accuracy of our method compared with  RA-SSL is in Table~\ref{tab:table1}.

\begin{figure}[b]
	\centering
	\includegraphics[width=0.8\columnwidth]{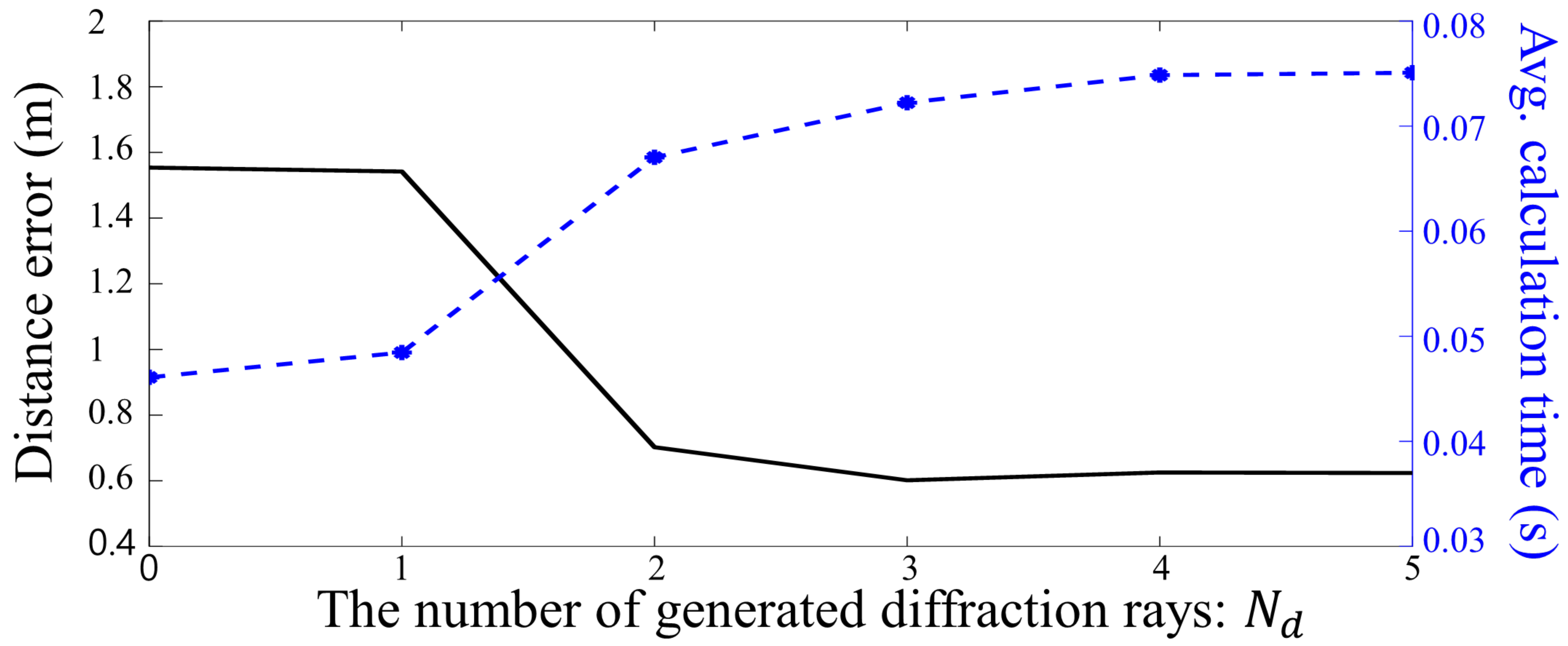}
	\vspace{-0.5em}
	\caption{
		This figure shows the average accuracy error and computation time for our method on
       an Intel i7 processor 6700, 
        as a function of $N_d$ the number of diffraction rays generated for simulating the edge diffraction.
       \Skip{
	When $N_d$ is 0 that means diffraction is not modeled and the performance corresponds to the RA-SSL algorithm~\cite{an2018reflection}.
		The average distance errors decreases and saturates after $N_d = 3$. 
		The restricted localization query time is $0.2$ seconds for a real-time. \YOON{fix the fig}}
	}
	\label{fig:analysis_diffraction}
\end{figure}

\paragraph{Analysis of diffraction rays.}
By modeling the diffraction effects, we increase the number of generated rays, resulting in a computational overhead.
As a result, we measure the average accuracy error and computation time as a function of $N_d$ the number of diffraction ray for simulating each edge diffraction.
As shown in Fig.~\ref{fig:analysis_diffraction}, 
the average accuracy error gradually decreases, but we found that when $N_d$ is in a range of 2 to 5, the accuracy is rather saturated.
Since we can accommodate to use up to $N_d = 5$ given our runtime computation
budget (0.2 s), we use $N_d=5$ across all the experiments.  In this case, the
average numbers of direct, reflection, and diffraction rays are $18$, $26$, and
$184$, respectively, in the case of the static source with clapping sound.
Also, the average running times for acoustic ray tracing and particle filter
are 0.09$ms$ and 72$ms$; our un-optimized  particle filter uses 100 particles
and computes weights of them against all the other rays.
When we are done on estimating the location within the time budget, we let our process to be in the idle state.

~\Skip{
\begin{table*}[t]
	\begin{center}
		\caption{Accuracies of the diffraction-aware SSL compared with RA-SSL}
		\label{tab:table1}
		\begin{tabular}{l|c|c|c|c|c|c} 
			{Source state} & \textbf{Stationary} & \textbf{Stationary} &  \textbf{Dynamic} &  \textbf{Dynamic(NLOS)} & \textbf{Total} &  \textbf{Total(NLOS)} \\
			{Sound} & Clapping & Male voice & Clapping & Clapping+male voice & Clapping+male voice \\
			\hline
			RA-SSL (avg. dist. error) & 1.4m & 1.12m & 1.15m & 1.83m & 1.22m & 1.45m \\			
			\hline
			Our method (avg. dist. error) & 0.6m & 0.82m & 0.7m & 0.95m & 0.7m & 0.79m\\
			\hline
			Improvement (\%)  & 130\% & 37\% & 64\% & 92\% & 77\% & 86\% \\
			\hline
		\end{tabular}
	\end{center}
\end{table*}
}

\begin{table}[t]
    \setlength\tabcolsep{6pt} 
	\begin{center}
		\caption{Summary of the accuracy of the different methods
		($*$: Only NLOS source)}
		\label{tab:table1}
		\begin{tabular}{l|c|c|c|c} 
			{Scenario} & \textbf{Stationary$*$} & \textbf{Stationary$*$} &  \textbf{Dynamic} &  \textbf{Dynamic$*$} \\
			{Sound} & Clapping & Male voice & Clapping & Clapping\\
			\hline
			RA-SSL & 1.4m & 1.12m & 1.15m & 1.83m \\		
			\hline
			Ours & 0.6m (130\%)& 0.82m(37\%) & 0.7m(64\%) & 0.95m(92\%)\\
			\hline
		\end{tabular}
	\end{center}
	\vspace{-1.5em}
\end{table}

\section{CONCLUSIONS \& FUTURE WORK}
\label{sec:6}

We have presented a novel diffraction-aware source localization algorithm. Our
approach can be used for localizing a NLOS source and models the diffraction
effects using the uniform theory of diffraction. We have combined our method
with indirect reflections and have tested our method in various scenarios with
static and moving sound sources with different sound signals.

While we have demonstrated benefits of our approach, it some limitations. The
UTD model is an approximate model and mainly designed for infinite wedges. As a
result, its accuracy may vary in different environment. We observed lower
accuracy for low-frequency sounds (male voice), mainly due to the diffuse
effect. Our implemented approach is limited to a single sound source in the
environment and does not model all the scattering effects. As part of future
work, we would like to address these problems.


{
	\bibliographystyle{ieee/ieee}
	\bibliography{Bib/robotics,Bib/soundSourceLocalization,Bib/Statistics,Bib/raytracing,Bib/soundPropagation}
}

\end{document}